\documentclass[unnumsec,webpdf,contemporary,large]{style}%
\theoremstyle{thmstyleone}%
\theoremstyle{thmstyletwo}%
\theoremstyle{thmstylethree}%

\definecolor{myred}{RGB}{150,84,84}
\definecolor{myblue}{RGB}{0,47,167}

\begin{document}
\appnotes{Published at Briefings in Bioinformatics}

\firstpage{1}

\title[BioGPT]{BioGPT: Generative Pre-trained Transformer for Biomedical Text Generation and Mining}

\author{Renqian Luo\ORCID{0000-0002-9062-3484}$^{1,\ast}$}
\author{Liai Sun$^{2}$}
\author{Yingce Xia\ORCID{0000-0001-9823-9033}$^{1,\ast}$}
\author{Tao Qin\ORCID{0000-0002-9095-0776}$^{1,\ast}$}
\author{Sheng Zhang\ORCID{0000-0003-3672-5436}$^{3}$}
\author{Hoifung Poon\ORCID{0000-0002-9067-0918}$^{3}$}
\author{Tie-Yan Liu$^{1}$}

\authormark{Renqian Luo et al.}

\address[1]{\orgname{Microsoft Research AI4Science}, \orgaddress{Beijing, China}}
\address[2]{\orgname{Peking University}, \orgaddress{Beijing, China}}
\address[3]{\orgname{Microsoft Research}, \orgaddress{Redmond, WA, USA}}

\corresp[$\ast$]{Corresponding authors: Renqian Luo, Microsoft Research AI4Science, E-mail: \href{email:renqianluo@microsoft.com}{renqianluo@microsoft.com}; Yingce Xia, Microsoft Research AI4Science, E-mail: \href{email:yinxia@microsoft.com}{yinxia@microsoft.com}; Tao Qin, Microsoft Research AI4Science, E-mail: \href{email:taoqin@microsoft.com}{taoqin@microsoft.com}}

\abstract{
Pre-trained language models have attracted increasing attention in the biomedical domain, inspired by their great success in the general natural language domain. Among the two main branches of pre-trained language models in the general language domain, i.e., BERT (and its variants) and GPT (and its variants), the first one has been extensively studied in the biomedical domain, such as BioBERT and PubMedBERT. While they have achieved great success on a variety of discriminative downstream biomedical tasks, the lack of generation ability constrains their application scope. In this paper, we propose BioGPT, a domain-specific generative Transformer language model pre-trained on large scale biomedical literature. We evaluate BioGPT on six biomedical NLP tasks and demonstrate that our model outperforms previous models on most tasks. Especially, we get $44.98\%$, $38.42\%$ and $40.76\%$ F1 score on BC5CDR, KD-DTI and DDI end-to-end relation extraction tasks respectively, and $78.2\%$ accuracy on PubMedQA, creating a new record. Our case study on text generation further demonstrates the advantage of BioGPT on biomedical literature to generate fluent descriptions for biomedical terms. Code is available at \url{https://github.com/microsoft/BioGPT}.
}

\maketitle

\section{Introduction}\label{intro}
Text mining and knowledge discovery from biomedical literature play important roles in drug discovery, clinical therapy, pathology research, etc. Typical tasks include recognizing named entities in the articles, mining the interaction between drugs and proteins/diseases/other drugs, answering questions given reference text, generating abstracts for given phrases/words, etc. People have accumulated large amounts of literature in the previous studies. For example, PubMed\footnote{\url{https://pubmed.ncbi.nlm.nih.gov}}, one of the most popular biomedical search engines, covers more than $30M$ articles and the number still rapidly increases every day as new discoveries are continuously coming out. Therefore, automatically mining the knowledge from literature becomes an urgent demand.  

Pre-training models have demonstrated their powerful capability in natural language processing~(NLP). On the GLUE benchmark, a widely used benchmark for natural language understanding, pre-training based methods outperform non-pre-training methods by a large margin~\cite{wang2018glue}\footnote{\url{https://gluebenchmark.com/leaderboard}}. There are two main kinds of pre-training models: (1) the BERT-like models~\cite{devlin2019bert,liu2019roberta,clark2019electra}, mainly for language understanding tasks; (2) the GPT-like models~\cite{radford2018improving,radford2019language,brown2020language}, mainly for language generation tasks.

These models are first pre-trained on large scale corpora collected from the Web via self-supervised learning task (e.g., masked language modeling for BERT, auto-regressive language modeling for GPT), and then fine-tuned on specific donwstream tasks. The BERT-like models are widely used in sequence classification and sequence labeling, where we need to encode the complete document. In comparison, the GPT-like models are often used in generation tasks (e.g., abstract generation, knowledge triplet generation).

By witnessing the success of pre-training in general NLP, people explore adapting these techniques into biomedical domain. However, directly applying these models to the biomedical domain leads to unsatisfactory performance due to domain shift~\cite{peng-etal-2019-transfer,pubmedbert}. A natural solution is to develop pre-training models on biomedical texts (e.g., PubMed). BioBERT~\cite{biobert} and PubMedBERT~\cite{pubmedbert}) are two representative BERT-like models pre-trained on biomedical domain, and they obtain superior performances than general pre-trained models on biomedical benchmarks. However, previous works mainly focus on BERT models which are more appropriate for understanding tasks, not generation tasks. In comparison, GPT models have demonstrated their abilities on generation tasks but demonstrate inferior performance when directly applying to the biomedical domain~\cite{moradi2021gpt, gutierrez2022thinking}.

In this work, we propose BioGPT, a domain-specific generative pre-trained Transformer language model for biomedical text generation and mining. BioGPT follows the Transformer language model backbone, and is pre-trained on $15M$ PubMed abstracts from scratch. We apply BioGPT to six biomedical NLP tasks: end-to-end relation extraction on BC5CDR~\cite{10.1093/database/baw068}, KD-DTI~\cite{hou2021discovering} and DDI~\cite{herrero2013ddi}, question answering on PubMedQA~\cite{jin2019pubmedqa}, document classification on HoC~\cite{baker2016automatic}, and text generation. To adapt to the downstream tasks, we carefully design and analyze the target sequence format and the prompt for better modeling the tasks. Experiments demonstrate that BioGPT achieves better performance compared to baseline methods and other well-performing methods across all the tasks.

\section{Related Work}\label{related}
\subsection{Pre-trained Language Models}
It has proven to be a very successful pattern in deep learning to pre-train models on large scale unlabeled data via careful designed self-supervision tasks and then transfer to downstream tasks by fine-tuning on them. Downstream tasks can benefit from the learned representations from the pre-trained models. BERT~\cite{devlin2019bert} is a bidirectional transformer based contextualized language model pre-trained on large scale text corpus English Wikipedia and BooksCorpus. It is pre-trained via carefully designed self-supervision tasks: \emph{masked language modeling~(MLM)} task where random word tokens of the input text are replaced by a special token [MASK] which is to be predicted by the model from the context, and the \emph{next sentence prediction~(NSP)} task where two sentences are to be predicted whether the second one is probable given the first one. The pre-trained BERT provides contextualized word representations that can be used by downstream tasks by just fine-tuning on the tasks and has achieved great success on various natural language understanding tasks. Subsequent works mainly focus on pre-training on larger-scale data and models~\cite{liu2019roberta} and advanced pre-training task~\cite{clark2019electra}. Though BERT and various biomedical BERT models have been successful in language understanding tasks and classification tasks, few efforts have been devoted to generative models. As BERT learns powerful word representations through the Transformer encoder model architecture in a bi-directional way, it limits its ability of generation.

Generative Pre-trained Transformer~(GPT)~\cite{radford2018improving} is proposed for language generation tasks via pre-training Transformer decoder model on large scale text corpus in a classical casual language modeling task where model learns to predict the next word token only dependent on the previous word tokens. Further, GPT-2~\cite{radford2019language} and GPT-3~\cite{brown2020language} with larger model size pre-trained on larger scale text corpus are proposed with remarkable performance in various downstream tasks (e.g., translation, summarization) including classification tasks (e.g., reading comprehension) even without fine-tuning (zero-shot) via appropriate prompts design. 

\subsection{Pre-trained Language Models in Biomedical Domain}
When applying to specific domain (e.g., biomedicine), BERT models pre-trained on general domain can be further improved if pre-trained on in-domain text data~\cite{peng-etal-2019-transfer,beltagy-etal-2019-scibert,biobert}. Specifically, \cite{biobert} and \cite{peng-etal-2019-transfer} start from the original pre-trained BERT model~\cite{devlin2019bert} that are pre-trained on general domain~(Wikipedia and BooksCorpus) and continue pre-training on biomedical literature. Specifically, \cite{biobert} continue pre-training using PubMed abstracts and PubMed Central full text articles and \cite{peng-etal-2019-transfer} continue pre-training on both PubMed text and clinical notes from MIMIC-III~\cite{johnson2016mimic}. As they are initialized from the original BERT that are pre-trained on general domain, they use the same vocabulary as the original BERT, which is quite different from the target biomedical domain. Instead of continue pre-training from the pre-trained BERT model, \cite{beltagy-etal-2019-scibert} pre-train the BERT model from scratch on large corpus of scientific literature (mainly biomedical and computer science literature) where the vocabulary is more suitable for science domain but still contains out-domain information for biomedicine. \cite{pubmedbert} propose that it is a better strategy to pre-train on domain-specific data from scratch where the vocabulary is more suitable for the biomedical domain. Consequently, they propose PubMedBERT which is pre-trained on $14M$ PubMed abstracts from scratch. Similarly, \cite{miolo2021electramed} pre-train on $28M$ data as in~\cite{peng-etal-2019-transfer} also from scratch, using the more advanced ELECTRA model. All these works have shown improved performance on plenty of biomedical literature language processing tasks compared to the original BERT pre-trained on general domain, while none of them is for biomedical generation tasks.

Noticing the powerful generation ability of GPT models, it is quite curious how GPT models perform on biomedical domain which is very different from general domain. However, recent works show that GPT models, even much more powerful GPT-3 model, perform poorly on biomedical tasks~\cite{moradi2021gpt, gutierrez2022thinking}. A previous work on pre-training GPT on biomedical literature is DARE~\cite{papanikolaou2020dare}. However, they pre-train GPT on very limited amount of data (only $0.5M$ PubMed abstracts) and use it only for data-augmentation for relation extraction task. A recent work on using GPT model is~\cite{agrawal2022large}, where they design converters for GPT-3~\cite{brown2020language} for several unconventional downstream clinical tasks.

\subsection{Downstream Tasks}
In this subsection, we introduce the downstream tasks we will work on. A summary of those tasks is in Table \ref{tbl:rel_tasks}. All these tasks can be formulated as text generation / mining tasks.

\begin{table*}[htbp]
    \caption{Summary of the downstream tasks}
    \label{tbl:rel_tasks}
    \small
    \centering
    \begin{tabular}{m{0.20\linewidth}|m{0.40\linewidth}|m{0.30\linewidth}}
        \toprule
        Task & Method & Dataset\\
        \midrule
        Relation Extraction & GLRE \cite{wang2020global},
        REBEL \cite{cabot2021rebel}, seq2rel \cite{giorgi2022sequence} & KD-DTI \cite{hou2021discovering}, BC5CDR \cite{10.1093/database/baw068}, DDI \cite{herrero2013ddi}\\
        \midrule
        Question Answering & QA-Net \cite{yu2018qanet}, LUKE \cite{yamada2020luke}, BERT \cite{devlin2019bert}, PubMedBERT \cite{pubmedbert}, BioELECTRA \cite{kanakarajan-etal-2021-bioelectra}, LinkBERT \cite{yasunaga2022linkbert} & PubMedQA \cite{jin2019pubmedqa}, BioASQ \cite{tsatsaronis2015overview,nentidis2019results}\\
        \midrule
        Document Classification &
        BERT \cite{devlin2019bert}, BlueBERT \cite{peng-etal-2019-transfer}, SciBERT \cite{beltagy-etal-2019-scibert}, SPECTER \cite{cohan2020specter}, PubMedBERT \cite{pubmedbert}, BioELECTRA \cite{kanakarajan-etal-2021-bioelectra}, LinkBERT \cite{yasunaga2022linkbert} & HoC \cite{baker2016automatic}, SciDocs \cite{cohan2020specter}\\
        \botrule
    \end{tabular}
\end{table*}

\subsubsection{Relation Extraction}
Relation extraction is a key task for biomedicine and life science research. Classical pipeline-based methods~\cite{zeng2014relation,zhou2016attention,wang2020global} resolve the task into several separate sub-tasks that require additional intermediate annotations and information which may suffer from the lack of intermediate annotated data and error accumulation. Joint extraction aims to jointly extract the entities and the relations between them from the text. Sequence labeling methods tackle the task by labeling the word tokens in the text with different tags to mark out all the entity mentions and then perform the relation classification between them via classifiers~\cite{sun2019joint,yuan2020relation,liu2021attention,wei2020novel}. Table filling methods formulate the task as a table constituted by the Cartesian product of itself and predicts the relations between the token pairs~\cite{fu2019graphrel,wang-etal-2020-tplinker,yan-etal-2021-partition}. These methods may suffer from error accumulation caused by previous tagging process and laborious intermediate annotations~(i.e., named entity recognition). Text generation methods reframe the task as a sequence-to-sequence learning task, by taking the text as the input sequence and the triplet as the target sequence and employing an encoder-decoder network to learn to generate the triplet from the text~\cite{zeng2018extracting,zhang2020minimize,hou2021discovering,cabot2021rebel,giorgi2022sequence}. However, many joint extraction methods still require additional entity information~\cite{wei2020novel,sui2020joint}. In this work, we focus on the end-to-end relation extraction, which formulates the task as an text generation task that takes only the text as the input and generates the relational triplets in an end-to-end way without additional intermediate annotations~\cite{cabot2021rebel,hou2021discovering,giorgi2022sequence}.

\subsubsection{Question Answering}
Question answering~(QA) is the task of answering questions given a context (reading comprehension). Typical methods predict a span in the source context as the answer text, or predicts a label (e.g., yes or no) for simpler tasks with predefined categorical answers \cite{yu2018qanet,hu2017reinforced,yamada2020luke}. \cite{pubmedbert,kanakarajan-etal-2021-bioelectra,yasunaga2022linkbert} mainly focus on the biomedical domain question answering task via pre-trained language models. Generative models \cite{radford2019language,brown2020language} directly generate the answer sequence or the label words.

\subsubsection{Document Classification}
Document classification is to classify a document into predefined categories (single label or multi label). Recent works on biomedical document classification also leverage large pre-trained language models for understanding the text and predicting the label \cite{peng-etal-2019-transfer,pubmedbert,kanakarajan-etal-2021-bioelectra,yasunaga2022linkbert}. Generative models \cite{radford2019language,brown2020language} generate the label words instead of predicting from the predefined set.

\section{Pre-training Method}
In this section, we describe our BioGPT from the perspective of dataset, vocabulary, and model.

\noindent{\bf Dataset}: Dataset is crucial for language model pre-training, in terms of amount, quality and domain. As Gu et al. \cite{pubmedbert} point, training only on in-domain data from scratch is important for specific domain. Therefore, we only consider in-domain text data and pre-train our model from scratch on the collected data. We collected all the PubMed items\footnote{\url{https://pubmed.ncbi.nlm.nih.gov}} that were updated before 2021 from the official site\footnote{\url{https://ftp.ncbi.nlm.nih.gov/pubmed/}} using the \texttt{wget} tool. We then filtered out all the empty items with only title but no abstract. We used the left $15M$ items (each with both title and abstract) as our pre-training dataset.

\noindent{\bf Vocabulary}: \cite{pubmedbert} also points that in-domain vocabulary is vital. Instead of using the vocabulary of GPT-2, we learn the vocabulary on our collected in-domain corpus. Specifically, we use byte pair encoding (BPE) \cite{sennrich2016neural} to segment the words in the corpus into word pieces and learn the vocabulary. We adopt the fastBPE\footnote{\url{https://github.com/glample/fastBPE}} implementation of BPE. The final learned vocabulary size is $42384$.

\noindent{\bf Model}: We adopt the GPT-2 model architecture~\cite{radford2019language} as the backbone of our BioGPT, which is a Transformer decoder~\cite{vaswani2017attention}. Currently we cannot follow the GPT-3 setting due to its extremely large model with 15 billion parameters. The core component of Transformer as well as our BioGPT is the multi-head attention. Given the input, three linear transformations are applied to produce the query $Q$, the key $K$ and the value $V$, and then the output is calculated as follows:
\begin{equation}
\begin{aligned}
& \texttt{Multihead}(Q,K,V)=\texttt{Concat}(head_1,head_2,\cdots,head_h)W, \\
& head_i=\texttt{softmax}\left(\frac{Q_iK_i^{T}}{\sqrt{d}}\right)V_i,
\end{aligned}
\label{eq:multihead_attn_highlevel}
\end{equation}
where (1) $h$ is the number of heads; (2) $Q$, $K$ and $V$ are equally split into $Q_i$, $K_i$ and $V_i$ along the feature dimension, $i\in\{1,2,\cdots,h\}$; (3) \texttt{Concat} denotes concatenating all inputs as a large tensor along the feature dimension; (4) $W$ is the parameter for the affine transformation. The output of multi-head attention layer is then fed into a feed-forward layer to construct a Transformer layer (or Transformer block). Practically, we adopt GPT-2$_\text{medium}$ as the backbone network which has 24 layers, 1024 hidden size and 16 attention heads resulting in $355M$ parameters in total, and our BioGPT has $347M$ parameters (the difference only comes from the different embedding size and output projection size caused by the different vocabulary size).

\noindent{\bf Training criteria}: BioGPT is trained via the standard language modeling task as the same as in \cite{radford2018improving,radford2019language}. Let $\mathcal{D}=\{x_i\}_i$ denote the collection of sequences, and sequence $x_i$ is made up of $n_i$ tokens, i.e., $x_i=(s_1,s_2,\cdots,s_{n_i})$. The training objective is to minimize the negative log-likelihood:
\begin{equation}
\min\;\;-\frac{1}{|\mathcal{D}|}\sum_{i=1}^{\vert\mathcal{D}\vert}\sum_{j=1}^{n_i}\log P(s_j|s_{j-1},s_{j-2},\cdots,s_1). 
\end{equation}

\section{Fine-tuning Method}
In this section, we introduce how to adapt the pre-trained BioGPT to downstream tasks: end-to-end relation extraction, question answering (QA) and document classification. The inputs of the tasks are all sequences, while they have different output formats.

To use BioGPT for these tasks, we need to convert the labels into sequences. In this way, the downstream task is consistent with the pre-training task in terms of the format.

Considering that BioGPT is pre-trained on massive natural language corpus, we convert the labels to sequences in natural language rather than the structured format using special tokens explored in other works~\cite{hou2021discovering,cabot2021rebel,giorgi2022sequence}. In this way, our reformed labels are semantically smoother than using special tokens. We will show the detailed implementation for each task and empirically verify the effectiveness of our method later.

\subsection{End-to-end Relation Extraction}
\noindent{\bf Task description}: Given a source sequence $x$, we need to find all triplets $\langle$head\_entity$_i$, tail\_entity$_i$, relation$_i\rangle_{i=1}^{N}$, that can be inferred from $x$. $N$ refers to the number of all possible triplets. Examples include extracting the drug-target-interaction, chemical-disease-relation and drug-drug-interaction. 

\noindent{\bf Method}: We convert the triplets into a simple natural language sequence with the same grammatical structures. We explore three forms in this paper:
\begin{enumerate}
    \item the ``subject verb object'' form (\texttt{svo}), where the entities correspond to the head entity, the relation and the tail entity in the triplet.
    \item the ``subject is the rel.noun of object'' form (\texttt{is-of}), where the ``rel.noun'' refers to the noun form of the relation.
    \item the `` the relation between subject and object is rel.noun'' form (\texttt{rel-is}).
\end{enumerate}

If there are multiple relational triplets for an input document, we sort them according to their order of appearance in the document and use semicolons to concatenated them together. 

Let us use a $\langle$drug, target, interaction$\rangle$ triplet as example. Suppose we would like to extract triplet $\langle$dextropropoxyphene (drug name),  mu-type opioid receptor (target name), inhibitor (relation)$\rangle$ from an input document. Then the \texttt{svo} representation is:
\begin{center}
    dextropropoxyphene inhibits mu-type opioid receptor.
\end{center}
The \texttt{is-of} form is:
\begin{center}
    dextropropoxyphene is the inhibitor of mu-type opioid receptor.
\end{center}
The \texttt{rel-is} form is:
\begin{center}
    the relation between dextropropoxyphene\\
    and mu-type opioid receptor is inhibitor.
\end{center}

The natural sentences can be converted back to triplets using regular expression. Users can also design customized formats depending on tasks.

\subsection{Question Answering}\label{sec:ft_qa}
\noindent{\bf Task description} Given a \emph{question}, a reference \emph{context} and an \emph{answer}, the goal is to answer the \emph{question} given the reference \emph{context}. The label is within the category of \emph{yes}, \emph{no}, or \emph{maybe}.

\noindent{\bf Method}: We pre-pend the description word ``question:'' and ``context:'' before the question and the context respectively and concatenate them together as the source sequence. Then for the target sequence, we generate it using the format ``the answer to the question given the context is \texttt{label}''. For example:
\begin{center}
\texttt{source}: question: \texttt{question text.} context: \texttt{context text.}\\
\texttt{target}: the answer to the question given the context is yes.
\end{center}

\subsection{Document Classification}\label{sec:ft_dc}
\noindent{\bf Task description} Given a document text, the goal is to classify the type of the document.

\noindent{\bf Method}: We generate the target sequence using the format ``the type of this document is \texttt{label}''. For example:
\begin{center}
the type of this document is genomic instability and mutation.
\end{center}

\subsection{Prompt-based Fine-tuning}
\begin{figure*}[htbp]
        \centering
        \includegraphics[width=0.8\linewidth]{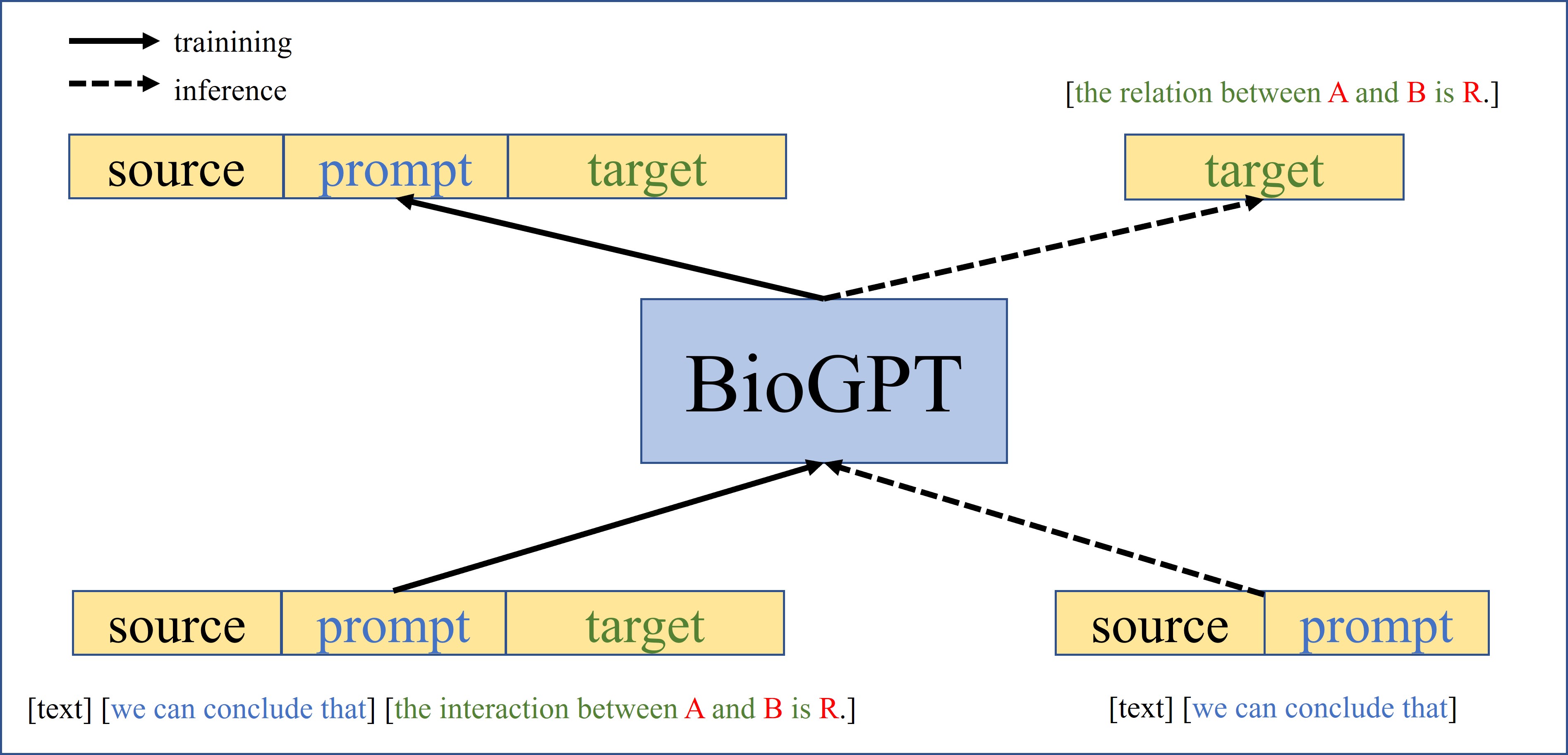}
        \caption{Framework of BioGPT when adapting to downstream tasks}
        \label{fig:framework}
\end{figure*}

We have formatted the labels to target sequences. The last question is, how do we use the source and the target to fine-tune and inference with BioGPT? A naive way is to concatenate the source and the target sequences together but is difficult for the model to generate during inference as it does not know what to generate for the specific task given the source text input.

Prompt is recently extensively explored in NLP~\cite{liu2021pre} to elicit the knowledge from a pre-trained language model. Prompt is to append task-specific instructions to the input for the model to better generate output that meets the demand of the task. GPT-3~\cite{brown2020language} uses hard prompts (manually designed discrete language phrases) to generate for different tasks. Though hard prompts can achieve satisfactory performance, designing task specific prompts is laborious and it is found that different prompts lead to different performance.

In this work, we mainly adopt soft prompts in prefix-tuning~\cite{li-liang-2021-prefix}, which leverage continuous embeddings (virtual tokens) to steer the pre-trained language model by directly appending several additional virtual tokens before the text as the prompts. Such continuous embeddings are randomly initialized and learned end-to-end on the downstream tasks to be task-specific. Different from~\cite{li-liang-2021-prefix}, we do not append the virtual tokens to the very beginning of the source input, but only before the target sequence (between the source and the target). Equipped with the prompt, our final sequence is constructed as $[\texttt{source};\texttt{prompt};\texttt{target}]$, as depicted in Fig.~\ref{fig:framework}. During the inference, we provide the source text and the prompt as the prefix for the language model to condition on and let the language model to generate the target output as in Fig.~\ref{fig:framework}.

\section{Experiments}
In this section, we pre-train our BioGPT and evaluate it on the following four biomedical NLP tasks across six datasets: end-to-end relation extraction on BC5CDR~\cite{10.1093/database/baw068}, KD-DTI~\cite{hou2021discovering} and DDI~\cite{herrero2013ddi}, question answering on PubMedQA~\cite{jin2019pubmedqa}, document classification on HOC~\cite{baker2016automatic}, and text generation on self-created dataset. We use fairseq~\cite{ott2019fairseq} as our code base for implementation. We adopt the GPT-2$_\text{medium}$ model configuration as our backbone model configuration. We perform BPE to learn to tokenize the corpus and construct the vocabulary instead of using the learned vocabulary from GPT-2 due to the domain gap between the biomedical domain and the general domain.

For pre-training, we pre-train BioGPT on 8 NVIDIA V100 GPUs for $200k$ steps, with 1024 tokens per GPU and 64 accumulated steps (i.e., the final batch size is $1024\times8\times64=524288$ tokens). We use Adam~\cite{kingma2015adam} as the optimizer with a peak learning rate of $2\times10^{-4}$ and 20000 warm-up steps. The learning rate follows an inverse square root decay schedule after reaching the peak as in~\cite{vaswani2017attention}.

All the fine-tuning experiments are conducted on a single NVIDIA V100 GPU, with a batch size of 1024 tokens and 32 accumulated steps.

During the inference, we adopt beam search with beam size=5 for the text generation task, and greedy search for all the other tasks. 

We make comparison to general domain GPT-2 for all the experiments. Specifically, we use the GPT-2$_\text{medium}$ model from the Hugging face library~\cite{wolf-etal-2020-transformers}\footnote{\url{https://huggingface.co/gpt2-medium}} which is the backbone network of our BioGPT.

\subsection{End-to-end Relation Extraction}
Relation extraction is an important task in information extraction. Here we target at the end-to-end relation extraction setting where the model takes the text as the input and directly generates the relational triplets. We mainly compare to REBEL~\cite{cabot2021rebel}, a recently proposed end-to-end triplet extraction approach based on sequence-to-sequence model, which employs BART pre-trained model~\cite{lewis-etal-2020-bart} as the backbone model, and further enhances it by pre-training on additional large relational triplet dataset created from Wikipedia as REBEL$_\text{pt}$.

\subsubsection{BC5CDR}
BC5CDR is a dataset for chemical-disease-relation extraction task introduced by~\cite{10.1093/database/baw068} which consists of 500/500/500 documents as the training/validation/test set. We fine-tune GPT-2$_\text{medium}$ and BioGPT for 100 epochs with a peak learning rate $10^{-5}$ and 100 warm-up steps. We use continuous embeddings with length=9 as prompts and the \texttt{rel-is} target sequence format. Since BC5CDR is a binary relation dataset where the entities are labeled if the relationship exists instead of a specific relation type, we use the pattern ``the relation between \texttt{head\_entity} and \texttt{tail\_entity} exists'' as the target sequence format. We average the checkpoints of the last 5 epochs for evaluation. We mainly measure and compare the micro-F1 score. We compare BioGPT to REBEL and seq2rel~\cite{giorgi2022sequence} where both methods are end-to-end relation extraction methods based on sequence-to-sequence modeling. We also compare with a pipeline-based extraction method, GLRE~\cite{wang2020global} which requires NER (named entity recognition) information as the intermediate annotations in the pipeline. Originally, GLRE uses the ground truth NER information. To make a fair comparison, we experiment with GLRE for two settings: 1) using ground-truth NER information during the training and using open-source NER tool during the inference (i.e., GLRE (gt+pred)) and 2) using open-source NER tool for both the training and the inference (i.e., GLRE (pred+pred)). We use the open-source NER tool\footnote{\url{https://huggingface.co/samrawal/bert-base-uncased_clinical-ner}} for the NER tagging. We try our best to run the baseline methods and evaluate them. 

\begin{table}[htbp]
    \caption{Results on BC5CDR chemical-disease-relation extraction task. 'gt+pred' means using ground truth NER information for training and using open-source NER tool to annotate NER for inference. 'pred+pred' means using open-source NER tool for both training and inference. '$\dagger$' means training on training and validation set.}
    \label{tbl:cdr}
    \small
    \centering
    \begin{tabular}{lccc}
       \toprule
       Model & Precision & Recall & F1 \\
       \midrule
       GLRE (gt+pred) & 34.82 & 18.29 & 23.99\\
       GLRE (pred+pred) & 23.00 & 4.88 & 8.05\\
       \midrule
       GPT-2~\cite{radford2019language} & 43.92 & 32.55 & 37.39 \\
       REBEL~\cite{cabot2021rebel} & 34.28 & 39.49  & 36.70 \\
       REBEL$_\text{pt}$~\cite{cabot2021rebel} & 40.94 & 21.20 & 27.94 \\
       seq2rel~\cite{giorgi2022sequence}$^{\dagger}$ & 43.5 & 37.5 & 40.2 \\
       \midrule
       BioGPT & 49.44 & 41.28 & \textbf{44.98}\\
       BioGPT$^{\dagger}$ & 49.52 & 43.25 & \textbf{46.17}\\
       \botrule
    \end{tabular}
\end{table}

From the results in Table~\ref{tbl:cdr}, we can see that BioGPT achieves the best result ($44.98\%$) among all the methods, with large improvements. We have several findings: 1) pipeline-based method GLRE significantly drops when using NER tagged by open-source tools instead of ground truth NER. However, this is often the common case in practical situation where the annotations for NER are lacked or expensive to collect. When applying open-source NER tools to some specific domains, errors occur and lead to inferior performance of relation extraction. 2) Compared to REBEL, BioGPT has a large gain with 8.28\% improvement. Notice that seq2rel~\cite{giorgi2022sequence} is trained on both the training set and validation set, while our BioGPT is only trained on the training set and still outperforms it with 4.78\% improvement. Moreover, when also trained on both the training set and the validation set, BioGPT further improves to 46.17\% with 5.97\% improvement against seq2rel~\cite{giorgi2022sequence}.

\subsubsection{KD-DTI}
KD-DTI is dataset for drug-target-interaction introduced by~\cite{hou2021discovering}, consisting of $12k$/$1k$/$1.3k$ documents as the train/validation/test set. We fine-tune GPT-2$_\text{medium}$ and BioGPT on the task for 30 epochs using Adam optimizer with a peak learning rate of $10^{-5}$ and 1000 warm-up steps. We use continuous embeddings with length=9 as prompts and the \texttt{rel-is} target sequence format for constructing the target sequence. We average the checkpoints of the last 5 epochs for evaluation. We mainly measure and compare the micro-F1 score and the results are listed in Table~\ref{tbl:kddti}. 

\begin{table}[htbp]
    \caption{Results on KD-DTI drug-target-interaction extraction task}
    \label{tbl:kddti}
    \small
    \centering
    \begin{tabular}{lccc}
        \toprule
        Model & Precision & Recall & F1 \\
        \midrule
        Transformer + PubMedBERT& \multirow{2}{*}{25.35} & \multirow{2}{*}{24.14} & \multirow{2}{*}{24.19} \\
        -attn~\cite{hou2021discovering} & & &\\
        GPT-2$_\text{medium}$ & 30.53 & 27.87 & 28.45 \\
        REBEL & 32.36 & 29.58  & 30.39 \\
        REBEL$_\text{pt}$ & 35.73 & 32.61 & 33.32 \\
        \midrule
        BioGPT & 40.00 & 39.72 & \textbf{38.42}\\
       \botrule
    \end{tabular}
\end{table}
We compare BioGPT with GPT-2$_\text{medium}$, Transformer + PubMedBERT-attn evaluated in~\cite{hou2021discovering} and REBEL. It can be shown that BioGPT achieves 38.42\% f1 score, with 14.23\%, 9.97\% and 8.03\% improvement compared to Transformer + PubMedBERT-attn, GPT-2$_\text{medium}$ and REBEL. Particularly, it surpasses REBEL$_\text{pt}$ by 5.1\% which is further pre-trained on large relation extraction dataset while BioGPT does not.

\subsubsection{DDI}
DDI extraction 2013 corpus is a dataset for drug-drug-interaction task introduced by~\cite{herrero2013ddi}, consisting of 792 texts selected from the DrugBank database
and other 233 Medline abstracts. We use the original dataset and use a train/validation/test split of 664/50/191 files. We fine-tune GPT-2$_\text{medium}$ and BioGPT for 100 epochs with a peak learning rate $10^{-4}$ and 500 warm-up steps. We also use continuous embeddings with length=9 as prompts and the \texttt{rel-is} target sequence format. The last 5 epochs are averaged for evaluation. The micro-F1 score is measured and compared.

\begin{table}[htbp]
    \caption{Results on DDI drug-drug-interaction extraction task}
    \label{tbl:ddi}
    \small
    \centering
    \begin{tabular}{lccc}
       \toprule
       Model & Precision & Recall & F1 \\
       \midrule
       GPT-2$_\text{medium}$ & 23.39 & 31.93 & 24.68 \\
       REBEL & 35.36 & 28.64  & 28.27 \\
       REBEL$_\text{pt}$ & 46.59 & 39.60 & \textbf{40.56} \\
       \midrule
       BioGPT & 41.70 & 44.75 & \textbf{40.76}\\
       \botrule
    \end{tabular}
\end{table}

The results are shown in Table~\ref{tbl:ddi} from which we can see that BioGPT achieves 40.76\% with 16.08\% and 12.49\% improvement against GPT-2$_\text{medium}$ and REBEL. It also surpasses REBEL$_\text{pt}$ which uses additional large relation extraction dataset for two-stage pre-training.

\subsection{Question Answering}
PubMedQA~\cite{jin2019pubmedqa} is a biomedical question answering dataset. Each sample is constructed from a PubMed abstract, containing a question, a reference context, a long answer, and a yes/no/maybe label which is the answer to the question. We use the original train/validation/test split with 450, 50 and 500 respectively, noted as PQA-L in \cite{jin2019pubmedqa} for evaluation. We also use the additional dataset noted as PQA-A and PQA-U in \cite{jin2019pubmedqa} for fine-tuning. We use the continuous embedding with length=9 as the soft prompt. We format the data into source sequence and target sequence as described before. We apply techniques such as two-stage fine-tuning~\cite{jin2019pubmedqa} and noisy labels to improve the performance. We measure and compare the classification accuracy of the reasoning required setting described in \cite{jin2019pubmedqa}.
\begin{table}[htbp]
    \caption{Results on PubMedQA question answering task}
    \label{tbl:pubm}
    \small
    \centering
    \begin{tabular}{lccc}
       \toprule
       Model & Accuracy \\
       \midrule
       PubMedBERT~\cite{pubmedbert} & 55.8 \\
       BioELECTRa~\cite{kanakarajan-etal-2021-bioelectra} & 64.2 \\
       BioLinkBERT$_\text{base}$~\cite{yasunaga2022linkbert} & 70.2 \\
       BioLinkBERT$_\text{large}$~\cite{yasunaga2022linkbert} & 72.2 \\
       \midrule
       BioGPT & \textbf{78.2}\\
       \botrule
    \end{tabular}
\end{table}

From the results in Table~\ref{tbl:pubm} we can see that BioGPT achieves 78.2$\%$ accuracy with 6.0$\%$ improvement over previous best performance obtained by BioLinkBERT~\cite{yasunaga2022linkbert}, achieving a new \emph{state-of-the-art} on this task.

\subsection{Document Classification}
HoC (the Hallmarks of Cancers corpus) consists of 1580 PubMed abstracts manually annotated at sentence level by experts with ten currently known hallmarks of cancer~\cite{baker2016automatic}. We follow the same training/test split as in~\cite{peng-etal-2019-transfer}. We use the continuous embedding with length=1 as the prompt and format the label into the target sequence as described before. We fine-tune GPT-2$_\text{medium}$ and BioGPT for 20000 steps with a peak learning rate $10^{-5}$ and 1000 warm-up steps. Micro-F1 score is measured and reported for comparison.
\begin{table}[htbp]
    \caption{Results on HoC document classification task}
    \label{tbl:hoc}
    \small
    \centering
    \begin{tabular}{lccc}
       \toprule
       Model & F1 \\
       \midrule
       BioBERT~\cite{biobert} & 81.54 \\
       PubMedBERT~\cite{pubmedbert} & 82.32 \\
       PubMedBERT$_\text{large}$ & 82.70\\
       BioLinkBERT$_\text{base}$ & 84.35 \\
       GPT-2$_\text{medium}$ & 81.84 \\
       \midrule
       BioGPT & \textbf{85.12}\\
       \botrule
    \end{tabular}
\end{table}

We can see from the results in Table~\ref{tbl:hoc} that BioGPT achieves $85.12\%$ accuracy with $3.28\%$ improvement over general domain GPT-2, and surpasses BioBERT, PubMedBERT and BioLinkBERT with $3.58\%$, $2.8\%$, $0.77\%$ improvements respectively.

\subsection{Text Generation}
GPT, GPT-2 and GPT-3 demonstrate remarkable text generation ability. Given words, phrases or simple sentences as prefix, they can continue to generate text that are syntactically correct and semantically smooth conditioning on the given text. We are also curious about the text generation ability of the pre-trained BioGPT in the biomedical domain, and how does general domain GPT-2 perform in the biomedical domain.

We evaluate the biomedical text generation ability of BioGPT and GPT-2$_\text{medium}$. Specially, we extract all the entities within the triplets from the KD-DTI test set (i.e., drugs and targets). Then for each drug/target name, we provide it to the language model as the prefix and let the model generate text conditioned on it. We then investigate whether the generated text is meaningful and fluent.

For this task, no objective evaluation metric is reported here. Instead, we provide a few examples here for demonstration.

\begin{table*}[htbp]
    \caption{Example of generated text}
    \label{tbl:gen_dt}
    \small
    \centering
    \begin{tabular}{m{0.10\linewidth}m{0.05\linewidth}m{0.75\linewidth}}
        \toprule
        Input & Model & Text \\
        \midrule
        Bicalutamide & GPT-2 & Bicalutamide, a synthetic bicalutamide derivative, was found to increase the cellular proliferation of C. elegans cells. This effect was not observed with the other synthetic bicalutamide derivatives. This is in line with our previous work which showed that bicalutamide increases the proliferation of C. elegans cells but not in the presence of synthetic bicalutamide.\\
        \cline{2-3}
        & BioGPT & Bicalutamide (Casodex) is an androgen receptor (AR) antagonist approved for the treatment of metastatic castration-resistant prostate cancer (mCRPC) in patients who have progressed on or are ineligible for docetaxel chemotherapy, as well as for the treatment of early-stage prostate cancer in men who have not progressed on or are ineligible for docetaxel chemotherapy, as well as for the treatment of metastatic castration-sensitive prostate cancer (mCSPC) in men who have not progressed on or are ineligible for docetaxel chemotherapy.\\
        \midrule
        Janus kinase 3 (JAK-3) & GPT-2 & Janus kinase 3 (JAK-3) mediates the conversion of glucose to glycogen in skeletal muscle. The increase in glucose uptake by skeletal muscle is believed to be associated with a reduction in muscle protein breakdown and with the release of ketone bodies, which contribute to the metabolism of glucose.\\
        \cline{2-3}
        & BioGPT & Janus kinase 3 (JAK-3) is a member of the Janus kinase (JAK) family of non-receptor tyrosine kinases and plays an important role in the regulation of cell proliferation, differentiation, survival, migration, and angiogenesis.\\
        \midrule
        Apricitabine & GPT-2 & Apricitabine, duloxetine, and citalopram) were investigated in patients with a history of treatment-resistant depression (n = 10). We found that treatment with the three drugs significantly decreased depressive symptoms, but not depressive symptoms over the next 12 weeks. We conclude that antidepressant drugs that reduce depressive symptoms and reduce depressive symptoms over time may be beneficial in patients with recurrent depression.\\
        \cline{2-3}
        & BioGPT & Apricitabine is an oral prodrug of 5-aza-2 '-deoxycytidine (5-aza-CdR), a DNA methyltransferase (DNMT) inhibitor, which has been approved by the US Food and Drug Administration (FDA) for the treatment of myelodysplastic syndrome (MDS) and acute myeloid leukemia (AML) in combination with low-dose cytarabine (Ara-C) and granulocyte colony-stimulating factor (G-CSF) for patients with intermediate-2 or high-risk MDS or AML.\\
        \midrule
        Xylazine & GPT-2 & Xylazine, the "active ingredient" in "bath salts" (see above) has been linked to numerous deaths, including those of former University of California, Berkeley, professor Andrew Anglin, who posted a video of himself having a bath filled with it.\\
        \cline{2-3}
        & BioGPT & Xylazine is an alpha 2-adrenoceptor agonist that has been used as a sedative and analgesic in veterinary medicine for many years, but its effects on the cardiovascular system have not been extensively studied in the dog, and its effects on the central nervous system (CNS) have not been well characterized in the dog, despite the fact that xylazine has been widely used as a sedative and analgesic in veterinary medicine for more than 30 years.\\
        \midrule
        Psoralen & GPT-2 & Psoralen, the city of the spiritualist, and had been given to him by God, who wished to show him the way of salvation, and to draw him back to the true life of Christ, and to bring him into communion with his fellow-man. \\
        \cline{2-3}
        & BioGPT & Psoralen photochemotherapy (PUVA) is a well-established treatment for psoriasis, but its use is limited by the risk of skin cancer, particularly squamous cell carcinoma (SCC) of the head and neck (H \& N), which is the most common site of skin cancer in the United States (US)\\
        \midrule
        CP-673451 & GPT-2 & CP-673451) and 4-5-6-3-1-2-3-4-6-7-8-9-10-11-12-13-14-15-16-17-18-19-20-21-22-23-24-25-26-27-28-29-30-31-32-33-34-35-36-37-38-39-40-41-42-43-44-45-46-47-48-49-50...\\
        \cline{2-3}
        & BioGPT & CP-673451 is a potent, selective, and orally active inhibitor of human neutrophil elastase (HNE) and human cathepsin G (CatG) with in vitro and in vivo anti-inflammatory activity in a variety of animal models of inflammation and in a model of acute lung injury (ALI) in the rat induced by intratracheal instillation of lipopolysaccharide (LPS) and tumor necrosis factor-alpha (TNF-alpha), a model of acute lung injury (ALI) in which neutrophils play an important role.\\
        \midrule
        BIIB-021 & GPT-2 & BIIB-021A-2-1.pdf. Wright, C. (2002). "The War on Terror: A Legal Theory of Terrorist Attacks," in L.E.R. Wright, S.R. Warrick, and J.H. Rothman (Eds.), The War on Terror: Essays in Honor of Michael T. Klare (New York: Oxford University Press), 143-57.\\
        \cline{2-3}
        & BioGPT & BIIB-021 is a novel, orally active, non-peptide bradykinin B2 receptor antagonist with potent and long-lasting anti-inflammatory activity in animal models of acute and chronic inflammation and in a rat model of adjuvant-induced arthritis (AIA), an animal model of rheumatoid arthritis (RA), and in a rat model of collagen-induced arthritis (CIA), an animal model of collagen-induced arthritis (CIA), in which arthritis is induced by immunization with bovine type II collagen (CII).\\
        \botrule
    \end{tabular}
\end{table*}

\begin{table*}[htbp]
    \caption{Generated text of keywords of interest}
    \label{tbl:gen_key}
    \small
    \centering
    \begin{tabular}{m{0.10\linewidth}m{0.05\linewidth}m{0.75\linewidth}}
        \toprule
        Input & Model & Text \\
        \midrule
        COVID-19 & GPT-2 & COVID-19-0067-7 Hip-hop-romance-detective-franchise-unveiled-in-coco-houston http://www.pinterest.com/pin/71178278499149422/ Black Widow: World War II's Greatest Widow? http://www.pinterest.com/pin/71177842668650112/\\
        \cline{2-3}
        & BioGPT & COVID-19 is a global pandemic caused by severe acute respiratory syndrome coronavirus 2 (SARS-CoV-2), the causative agent of coronavirus disease 2019 (COVID-19), which has spread to more than 200 countries and territories, including the United States (US), Canada, Australia, New Zealand, the United Kingdom (UK), and the United States of America (USA), as of March 11, 2020, with more than 800,000 confirmed cases and more than 800,000 deaths.\\
        \midrule
        SARS-CoV-2 & GPT-2 & SARS-CoV-2-CKM-T1-1/pdf/T1/SARS-CoV-2-CKM-T1-1.pdf Cancer http://www.cancer.gov/cancer
        topics/programs/tox/tox-programs.html Other Research http://www.cdc.gov/cancer/cancer/index.html\\
        \cline{2-3}
        & BioGPT & SARS-CoV-2 is the causative agent of COVID-19, a severe acute respiratory syndrome (SARS) that has infected more than 390,000 people worldwide and killed more than 250,000 people.\\
        \midrule
        The treatment of COVID-19 & GPT-2 & The treatment of COVID-19 is similar to that of COVID-20 and COVID-22, in which COVID-19 is treated with a variety of ionic, dopaminergic and serotonergic agents. In addition to its effects on COVID-19, the use of COVID-19 has been suggested to exert effects on the function of other dopamine neurons, including dopamine D2 receptors, D3 receptors and serotonin transporters. These effects include attenuation of the increase in dopamine release induced by the central nervous system stimulant drug amphetamine and its metabolites.\\
        \cline{2-3}
        & BioGPT & The treatment of COVID-19 with remdesivir (remdesivir for COVID-19) has been approved by the US Food and Drug Administration (FDA) for the treatment of severe acute respiratory syndrome coronavirus 2 (SARS-CoV-2) infection in patients with confirmed SARS-CoV-2 infection and is currently being evaluated in clinical trials for the treatment of COVID-19 in patients with confirmed SARS-CoV-2 infection, as well as in patients with suspected or confirmed SARS-CoV-2 infection.\\
        \midrule
        The drug that can treat COVID-19 is & GPT-2 & The drug that can treat COVID-19 is not effective enough to cure patients with this rare genetic disease, but it could help patients who have it. Researchers say this finding could mean that a drug could be developed that would help patients with this rare genetic disorder. The study, published in the journal Molecular Psychiatry, also looked at how a drug might be used to treat patients with COPD, which causes chronic pain and lung damage. Researchers say the study could also lead to new drugs that treat COPD.\\
        \cline{2-3}
        & BioGPT & The drug that can treat COVID-19 is hydroxychloroquine (HCQ), which has been shown to inhibit the replication of the SARS-CoV-2 virus in cell culture and in animal models of COVID-19, and has been approved by the US Food and Drug Administration (FDA) for the treatment of patients with COVID-19 in March 2020, and by the European Medicines Agency (EMA) for the treatment of patients with COVID-19 in April 2020, and by the European Medicines Agency (EMA) for the treatment of patients with COVID-19 in May 2020.\\
        \midrule
        Omicron variants of SARS-CoV-2 & GPT-2 & Omicron variants of SARS-CoV-2 have been detected in the blood of patients who contracted SARS-CoV-2 and survived it, but it is unclear whether these variants were caused by SARS-CoV-2 infection or if the SARS-CoV-2 variant in the blood was due to a misidentification of SARS-CoV-2 from other SARS strains. We found that a polymorphism in the gene encoding the SARS-CoV-2-specific viral surface protein was associated with SARS-CoV-2 infection in a cohort of patients with SARS-CoV-2 infection who had an active SARS infection, suggesting that SARS-CoV-2 may be able to infect the host during an active infection.\\
        \cline{2-3}
        & BioGPT & Omicron variants of SARS-CoV-2 have been isolated from patients with severe acute respiratory syndrome (SARS) and have been shown to be highly pathogenic in mice and ferrets, suggesting that they may play a role in the pathogenesis of SARS-CoV-2 infection and the development of severe disease in patients with SARS-CoV-2 infection.\\
        \botrule
    \end{tabular}
\end{table*}

From the results in Table~\ref{tbl:gen_dt}, we can see that: (1) Given relatively common names as input, for example in the first two cases (i.e., Bicalutamide and JAK-3), GPT-2 can generate meaningful and fluent text that is related to the word and biomedicine, while BioGPT generates more specific and professional descriptions. (2) When given some uncommon names (e.g., in the Apricitabine and Xylazine cases), GPT-2 cannot generate meaningful descriptions while BioGPT still generates specific descriptions. Especially in the Apricitabine case, GPT-2 seems to generate a piece of text that comes from a specific scientific paper while BioGPT generates more general description. (3) When given some very uncommon and domain specific names that even lose semantic information from their surface names (e.g., Psoralen, CP-673451 and BIIB-021), GPT-2 trained on general completely failed to generate any informative text. Given Psoralen, GPT-2 treats it as a city name and generates some text though fluent but unrelated to the given name. Given CP-673451, GPT-2 even begins to count numbers. Given BIIB-021, GPT-2 treats it as a name of a pdf document. For these types, BioGPT is still able to generate text that describes the names or is highly related to them.

Besides these samples, we also manually input several keywords or phrases that are of interest (e.g., COVID-19 related terms) and see what GPT-2 and our BioGPT generate. The results are listed in Table~\ref{tbl:gen_key}, where we input many COVID-19 related key words/phrases as the prefix for the language model to condition on. We can see that GPT-2 treats the term ``COVID-19'' and ``SARS-CoV-2'' as some codes within a link or file name rather the entities we care about while BioGPT can generate clear descriptions. More interestingly, when prompting ``The drug that can treat COVID-19 is'', BioGPT is able to answer it with the drug ``hydroxychloroquine'' which is indeed noticed at MedlinePlus\footnote{\url{https://medlineplus.gov/druginfo/meds/a601240.html}}. Notice that GPT-2 is pre-trained on the corpus before COVID-19 while BioGPT is pre-trained on the corpus before 2021 that contains COVID-19 information, therefore it is not surprising that BioGPT performs much better than GPT-2 on COIVD-19 related key words in Table~\ref{tbl:gen_key}. However, in the last example in Table~\ref{tbl:gen_key}, both models do not have any knowledge of the Omicron variants of SARS-CoV-2 which appear in the late 2021, while BioGPT still generates more fluent and relevant text compared to GPT-2.

Overall, we can see that BioGPT pre-trained on in-domain biomedical literature from scratch performs better than general domain GPT-2 across various biomedical NLP tasks, and performs better than most previous methods on respective tasks, achieving \emph{state-of-the-art} on four out of six tasks.

\section{Ablation Study}
In this section, we conduct ablation study on the prompt design and the target sequence format of the label.

\begin{table*}[htbp]
    \caption{Results on KD-DTI with different target formats}
    \label{tbl:target}
    \small
    \centering
    \begin{tabular}{lccc}
        \toprule
        Target format & Precision & Recall & F1 \\
        \midrule
        \texttt{\textcolor{myred}{<head>}} \textcolor{myblue}{head\_entity} \texttt{\textcolor{myred}{<tail>}} \textcolor{myblue}{tail\_entity} \texttt{\textcolor{myred}{<relation>}} \textcolor{myblue}{relation}
        & 38.21 & 40.21 & 37.32 \\
        \texttt{svo} (\textcolor{myblue}{head\_entity} \textcolor{myblue}{relation} \textcolor{myblue}{tail\_entity})
        & 37.95 & 37.77 & 36.57 \\
        \texttt{is-of} (\textcolor{myblue}{head\_entity} is the \textcolor{myblue}{relation} of \textcolor{myblue}{tail\_entity})
        & 39.37 & 39.11 & 37.77 \\
        \texttt{rel-is} (the relation between \textcolor{myblue}{head\_entity} and \textcolor{myblue}{tail\_entity} is \textcolor{myblue}{relation})
        & 38.93 & 40.70 & 38.38\\ 
       \botrule
    \end{tabular}
\end{table*}

\subsection{Target Sequence Format}
Previous works~\cite{cabot2021rebel,phan2021scifive,giorgi2022sequence,hou2021discovering} directly format the labels into structured formats using special tokens. Taking the triplet generation task as an example, in REBEL~\cite{cabot2021rebel}, the triplets are represented by:
\begin{center}
\texttt{\textcolor{myred}{<triplet>}} \textcolor{myblue}{head\_entity$_1$} \texttt{\textcolor{myred}{<subj>}} \textcolor{myblue}{tail\_entity$_1$} \texttt{\textcolor{myred}{<obj>}} \textcolor{myblue}{relation$_1$}\\
\texttt{\textcolor{myred}{<triplet>}} \textcolor{myblue}{head\_entity$_2$} \texttt{\textcolor{myred}{<subj>}} \textcolor{myblue}{tail\_entity$_2$} \texttt{\textcolor{myred}{<obj>}} \textcolor{myblue}{relation$_2$} $\cdots$,
\end{center}
where \texttt{<triplet>}, \texttt{<subj>} and \texttt{<obj>} are special tokens to represent the start of the head entity, the tail entity and the relation.~\cite{cabot2021rebel,hou2021discovering,giorgi2022sequence} use similar method to process the targets.

Although these methods achieved promising results in their tasks respectively, such formulation pattern is not the best choice for BioGPT. Previous works use an encoder-decoder framework, where two separated modules are leveraged to process the input (by the encoder) and generate the answers (by the decoder). The two modules can be trained to fit the two different types of sequences (natural language sequence v.s. structured sequence). 

In contrast, in BioGPT, we use a unified module to encode context and generate answers. Intuitively, it is better to maintain the format consistency between the inputs and answers. Consequently, instead of the structured target format with special tokens as in previous works, we format the label within a natural language sentence for the language model to smoothly learn and generate. However, there are also various patterns that can be used to construct the target sentence. We explore several target sequence formats, including the structured format, on the KD-DTI dataset for end-to-end relation extraction task. We fix the prompt to continuous embeddings with length=9. From the results in Table~\ref{tbl:target} we can see that the formats in natural language perform better than structured format, and that the \texttt{rel-is} format performs the best among all the formats in terms of F1 which provides a more semantically smooth and clear description. We also conduct experiments on BC5CDR and DDI to further compare the structure format and the \texttt{rel-is} format. The F1 scores of the structure format on BC5CDR and DDI are $42.85$ and $38.60$, while those two scores with \texttt{rel-is} format are $44.98$ and $40.76$, which further verify our conclusion.

\subsection{Prompt Design}
\begin{table}[htbp]
    \caption{Results on KD-DTI with different prompts}
    \label{tbl:prompt}
    \small
    \centering
    \begin{tabular}{lccc}
       \toprule
       Prompts & Precision & Recall & F1 \\
       \midrule
       we have that & 38.55 & 38.37  & 36.95  \\
       in conclusion, & 39.03 & 39.45 & 37.76 \\
       we can conclude that & 39.56 & 39.88 & 38.16 \\
       continuous embeddings (length=1) & 39.50 & 39.71 &  38.06 \\
       continuous embeddings (length=5) & 39.57 & 39.63 & 38.09 \\
       continuous embeddings (length=9) & 38.93 & 40.70 & 38.38 \\
       continuous embeddings (length=13) & 39.48 & 39.17 & 38.60 \\
       continuous embeddings (length=17) & 39.82 & 39.60 & 38.28 \\
       \botrule
    \end{tabular}
\end{table}
We conduct experiment with manually designed hard prompts and continuous embedding soft prompts on the KD-DTI extraction task. We fix the target format to the \texttt{rel-is} format (i.e., "the relation between \texttt{head\_entity} and \texttt{tail\_entity} is \texttt{relation}"). From the results in Table~\ref{tbl:prompt} we can see that the best performing prompt is continuous embeddings with length of 13 virtual tokens. Moreover, we have several observations: (1) Different manually designed hard prompts result in different performance and more instructive and informative prompt (e.g., ``we can conclude that'') achieve better performance. (2) Generally, continuous embedding soft prompts perform better than manually designed hard prompts. (3) The performance of the continuous embedding soft prompts are roughly irrelevant to the length. In our previous experiments, we empirically choose length=9 according to the performance on validation set.

\section{Conclusion}
In this work, we proposed BioGPT, a generative pre-trained Transformer language model for biomedical text generation and mining. We adopted GPT-2 as our backbone model and pre-trained on $15M$ PubMed abstracts corpus. We carefully designed and investigated the prompt and the target sequence format when applying pre-trained BioGPT to downstream tasks. We applied the pre-trained BioGPT to biomedical NLP tasks: end-to-end relation extraction task, question answering task, document classification task and text generation task. BioGPT achieves SOTA results on three end-to-end relation extraction tasks and one question answering task. It also demonstrates better biomedical text generation ability compared to GPT-2 on the text generation task. For future work, we plan to train larger scale BioGPT on larger scale biomedical data and apply to more downstream tasks.\\

\noindent\fbox{
\begin{minipage}{0.90\columnwidth}
\textbf{Key Points}\\
    Our contributions are summarized as follows:
    \begin{itemize}
    \item We propose BioGPT, a generative pre-trained Transformer language model on biomedical domain. BioGPT can be used for biomedical literature text generation and mining.
    \item BioGPT achieves state-of-the-art results on four benchmarks: BC5CDR, KD-DTI and DDI end-to-end relation extraction task, and PubMedQA question answering task. We also demonstrate the capability of biomedical text generation of BioGPT compared to standard GPT trained on general domain.
    \item We study the prompt design and the target sequence design when applying BioGPT to downstream tasks and find that target sequence with natural language semantics are better than structured prompts explored in previous works.
    \end{itemize}
\end{minipage}
}

\begin{appendices}
\section{Scaling to Larger Size}
We also scaled our model to larger size. We built BioGPT-Large, based on the GPT-2 XL architecture (the largest version of GPT-2), with 1.5B model parameters. We fine-tune and evaluate its performance on the downstream tasks, as shown in Table \ref{tbl:pubmfull}.
\begin{table}[htbp]
    \caption{Performance of BioGPT-Large fine-tuned on downstream tasks}
    \label{tbl:pubmfull}
    \small
    \centering
    \begin{tabular}{lcc}
       \toprule
       Task & Metric & Performance \\
       \midrule
       BC5CDR & F1 & 50.12\\
       KD-DTI & F1 & 38.39\\
       DDI & F1 & 44.89\\
       PubMedQA & Accuracy & 81.0 \\
       HoC & F1 & 84.40 \\
       
       \botrule
    \end{tabular}
\end{table}

\end{appendices}

\bibliographystyle{unsrt}
\bibliography{main}


\end{document}